\documentclass[journal]{./IEEEtran}
\usepackage{natbib}
\usepackage{amsmath}

\usepackage[]{algorithm2e}
\usepackage{listings}
\usepackage{url}
\usepackage{proof}
\usepackage{tikz}
\usepackage{hyperref}

\usepackage{graphicx}
 
\usetikzlibrary{trees}
\usetikzlibrary{arrows}

\DeclareMathOperator*{\argmax}{arg\,max}

\usetikzlibrary{bayesnet}

\usepackage{amssymb}
\usepackage{amsthm}
\usepackage{stfloats}

\begin{document}
%
\title{Neural Decision Trees}

\author{
    \IEEEauthorblockN{Randall Balestriero\IEEEauthorrefmark{1}}
    \\
    \IEEEauthorrefmark{1} Electrical and Computer Engineering Department, Rice University, Houston TX, USA
    \\
    \IEEEauthorblockA{\IEEEauthorrefmark{1} 
    randallbalestriero@gmail.com}
}

\maketitle

\begin{abstract}
In this paper we propose a synergistic melting of neural networks and decision trees (DT) we call neural decision trees (NDT). NDT is an architecture a la decision tree where each splitting node is an independent multilayer perceptron allowing oblique decision functions or arbritrary nonlinear decision function if more than one layer is used. This way, each MLP can be seen as a node of the tree.
We then show that with the weight sharing asumption among those units, we end up with a Hashing Neural Network (HNN) which is a multilayer perceptron with sigmoid activation function for the last layer as opposed to the standard softmax. The output units then jointly represent the probability to be in a particular region. The proposed framework allows for global optimization as opposed to greedy in DT and differentiability w.r.t. all parameters and the input, allowing easy integration in any learnable pipeline, for example after CNNs for computer vision tasks.
We also demonstrate the modeling power of HNN allowing to learn union of disjoint regions for final clustering or classification making it more general and powerful than standard softmax MLP requiring linear separability thus reducing the need on the inner layer to perform complex data transformations.
We finally show experiments for supervised, semi-suppervised and unsupervised tasks and compare results with standard DTs and MLPs.
\end{abstract}

\IEEEpeerreviewmaketitle

\section{Introduction and Decision Tree Review}
\subsection{Motivations}
\IEEEPARstart{H}{ashing}. A hash function $H:X\rightarrow Y$ is a many-to-one mapping from a possibly infinite set to a finite set \cite{harrison1971implementation} where we usually have $X \subset \mathbb{R}^d$ and $Y\subset \{1,...,C\}$. As a result, the following function
\begin{equation}
H(x)=\argmax_y p(y|x),
\end{equation}
typical to discriminative models \cite{jordan2002discriminative}, can be seen as a hashing function that has been optimized or learned given some dataset of $N$ independent observations $\{(x_i,y_i)_{i=1}^N\}$ of input data $x_i$ and corresponding label $y_i$. Decision trees (DT) are a special kind of discriminative model aiming at breaking up a complex decision into a union of simple binary decision a.k.a splitting nodes \cite{safavian1990survey}. In order to do so, DT learning involves a sequential top-down partitioning of the input space $X$ into sub-regions $\Omega_k$ satisfying
\begin{align}
&\cup_k \Omega_k = X, \\
&\Omega_k\cap \Omega_l = \emptyset, \forall k \not = l.
\end{align}
This partition is done so that the label distribution of the points w.r.t each region has minimal entropy. In particular, the optimum is obtained when all the points lying in a sub-region belong to the same class and this for all the sub-regions. Once trained, a new observation $x$ is classified by first finding in which region $\Omega(x)$ it belongs to and by predicting the label specific to this region according to
\begin{equation}
H(x)=mode(\{y_i|x_i \in \Omega(x)\}),
\end{equation}
for classification problems and
\begin{equation}
H(x)=mean(\{y_i|x_i \in \Omega(x)\}),
\end{equation}
for regression problems.
What brings DT among the most powerful discriminative technique for non cognitive tasks lies in the fact that the number of sub-regions $\Omega_k$ of $X$ grows exponentially w.r.t their depth. This milestone is the core of the developed Hashing Neural Network (HNN) coupled with the acute modeling capacity of deep neural networks. We now describe briefly the standard univariate and multivariate decision trees, their advantages and drawbacks as well as motivations to extend them in a more unified and differentiable framework. In fact, as Quinlan said \cite{quinlan1994comparing}, decision trees are better when discrimination has to be done through a sequential testing of different attributes whereas ANN are good when knowledge of a simultaneous combination of many attributes is required, trying to get the best of both worlds seems natural.
\subsection{Univariate and Oblique Decision Trees}
Decision trees lead to a recursive space partitioning of the input space $X$ through a cascade of simple tests \cite{kuncheva2004combining}.
In the case of univariate or monothetic DT, the local test for each splitting node is done by looking at one attribute $att$ of $x\in X$ and comparing the taken values w.r.t. a threshold value $b$. If we have $x(att)<b$ then the observation is passed to the left child where another test is performed. This process is repeated until reaching a leaf in which case a prediction can be done for $x$. 
There is an intuition behind the sequence of simple tests performed by the tree to classify an object which is particularly useful in botany, zoology and medical diagnosis.
By looking at all the leaves, one can see that they partition the input space into a set of axis-aligned regions. 
Since it has been proven that growing the optimal tree is NP-complete \cite{hyafil1976constructing} standard DT induction performs a greedy optimization by learning the best attribute and threshold for each split sequentially \cite{quinlan1986induction} unless everything is picked at random s.a. in Extremely Randomized DT \cite{geurts2006extremely}.
It is usually built in a top-down fashion but bottom-up and hybrid algorithms also exist.
If a stopping criteria has been used during the growing phase it is called prepruning.
Typical stopping criteria or prepruning can involve a validation set, criteria on the impurity reduction or on the number of examples reaching a node. Finally, hypothesis testing using a Chi-Square test \cite{goulden1939methods} to see if the class distribution of the children is different than the parent can also be used.
Early stopping can be detrimental by stopping the exploration, known as the horizon effect \cite{duda2012pattern}, a DT can instead be fully developed and then postpruned using some heuristics which is a way to regularize the splits. In fact, in the case of a fully developed DT and if there is not two identical objects $x_i, x_j$ in the dataset with different class label, the decision tree can learn the dataset, leading to zero re-substitution error but making them instable. In fact, it can memorize the training set and thus a small change in the input would lead to a completely different fitted tree.
The main limitation of univariate DT resides in the axis-aligned splits. This inherently implies that the performance of DT is not invariant to the rotation of the input space and for cognitive tasks, tests that are done only on one attribute at a time lead extremely poor results if some hand-crafted features are not provided. As a result, oblique decision trees have been developed for which a test is now done on a linear combination of the attributes of $x$. Since the splitting is still not differentiable, optimizing the cutting value and the weight vector $w$ is usually done with Genetic Algorithms (GA)\cite{cantu2003inducing}. Finally, unsupervised pre-processing has also been developed with Random Projection Trees  \cite{dasgupta2008random,blaser2015random} and PCA Trees \cite{verma2009spatial,sproull1991refinements} to avoid the rotation problems.
A partitioning example is presented in Fig. \ref{random} for a univariate and oblique tree.
\begin{figure}[t!]
\begin{center}
\includegraphics[width=3.5in]{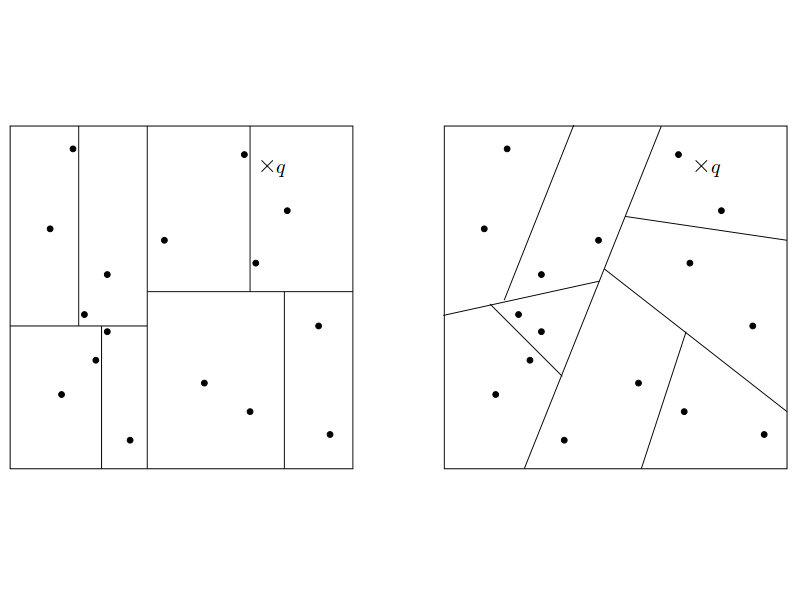}
\caption{Random Projection Tree example on $2D$.}
\end{center}
\end{figure}\label{random}
The contributions of the paper are: learning arbitrary decision boundaries as opposed to axis-aligned or linear through a global optimization framework as opposed to greedy optimization via differentiable splitting nodes. The derivation of the HNN allowing the use of deep neural networks with a new deep hashing layer which is related to Locality Sensitive Hashing (LSH). Finally, the use of the developed HNN for supervised and unsupervised problems as well as classification and regression tasks.
\section{Neural Decision Trees}
We first introduce the neural decision tree (NDT) which is a soften version of decision trees allowing finely learned arbitrary decision surfaces for each node split and a global optimization framework. We first review briefly supervised LSH as the neural decision tree is a particular instance of a LSH framework.

\subsection{Locality Sensitive Hashing}
Locality Sensitive Hashing \cite{gionis1999similarity,charikar2002similarity} aims at mapping similar inputs to the same hash value. In the case of trees, the hash value corresponds to the reached leaf. Learning this kind of function in a supervised manner has been studied \cite{liu2012supervised}.
For example in \cite{xia2014supervised} the similarity matrix induced by the labels is factorized into $H^TH$ and the features $H$ are learned through a CNN.
In \cite{salakhutdinov2009semantic} a deep autoencoder is learned in an unsupervised manner and the latent representation is then used for LSH. This last framework will be a special case of our unsupervised HNN with the main difference that the autoencoder will not just be trained to reconstruct the input but also provide a meaningful clustered latent space representation.
\subsection{Model}
\begin{figure}[t!]
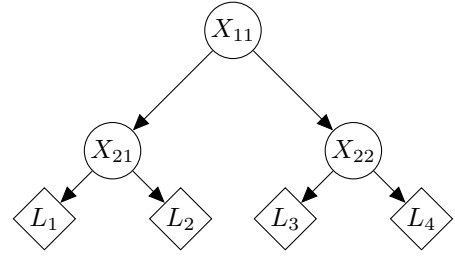

  \begin{center}
  \tikz{
        \node[latent]                    (x1) {$X_{11}$}; %
        \node[latent,below right=1.5 of x1]       (x22) {$X_{22}$};
        \node[latent,below left=1.5 of x1]       (x21) {$X_{21}$};
		\node[det,below right=0.6 of x21]       (x31) {$L_2$};
        \node[det,below left=0.6 of x21]       (x32) {$L_1$};
		\node[det,below right=0.6 of x22]       (x34) {$L_4$};
        \node[det,below left=0.6 of x22]       (x33) {$L_3$};              
        \edge {x1}{x21,x22}; 
        \edge {x21}{x31,x32};
        \edge {x22}{x33,x34};
      } 
  \end{center}\caption{Simple Tree}\label{tree}
\end{figure}

The main change we perform on a decision tree to make it differentiable is to replace the splitting function which can be seen as an indicator function into a sigmoid function
\begin{equation}
\Phi(x)=\frac{1}{1+e^{-x}}.
\end{equation}
We now interpret for each node $i,j$ the output of $\Phi(x)_{i,j}$ as the probability that the instance $x$ goes to the left child of the node, note that this is a generalized version of the node change suggested in \cite{laptev2014convolutional}. For example, looking at the tree representation \ref{tree},
we have
\begin{equation}
P(x\in L_1)=P(X_1=\text{left}|x)P(X_{22}=\text{left}|X_1=\text{left},x).
\end{equation}
In fact, if we denote the attributes and threshold value for each node $X_{i,j}$ in the tree as $att_{X_{i,j}}$ and $b_{X_{i,j}}$, we can denote the probability of a point to be passed to the left child through the use of a sigmoid function as
\begin{align}
&P(X_{i,j}=\text{left}|x,X_{\pi_{i,j}})=1_{x(att_{X_1})<b_{X_1}} && \text{for DT},\\
&P(X_{i,j}=\text{left}|x,X_{\pi_{i,j}})=\Phi_{i,j}(x) &&\text{for NDT},
\end{align}
where $\Phi_{i,j}(x)$ denotes the sigmoid at node $i,j$ containing the hyper-parameters and $\pi_{i,j}$ the decisions made by the parent of node $i,j$. More generally, we can apply the sigmoid after application of a general transformation of the data $G:\mathbb{R^D}\Rightarrow \mathbb{R}$ such as a ANN as
\begin{equation}
P(X_{i,j}=\text{left}|x,X_{\pi_{i,j}},G_{i,j})=\Phi_{i,j}(G_{i,j}(x)).
\end{equation}
We now denote this probability directly as $\Phi_{i,j}(x)$ including the function $G_{i,j}$ and the parent decisions.
\subsection{Local and Global Loss}
We now introduce some notations in order to derive the overall loss function of the tree. We will derive the binary classification case only since the general case will be developed in the HNN section. We thus restrain $y\in \{0,1\}$. 
In the case where each node is optimized in a top-down fashion we will first introduce some notation and derive the soft version of the usual DT splitting criteria. First, we have for each node $i,j$ which is not a leaf
\begin{align}
&P = \sum_{n=1}^ny_n,\\
&N = \sum_{n=1}^n(1-y_n),\\
&n_{left}  = \sum_{n=1}^n\Phi(x_n),\\
&n_{right} = P+N-n_{left},\\
&P_{left}  = \sum_{n=1}^n\Phi(x_n)y_n,\\
&N_{left}  = n_{left}-P_{left},\\
&P_{right} = P-P_{left},\\
&N_{right} = N-N_{left},
\end{align}
where these quantities represent respectively the number of positive and null observations, the observations going to the left and right node, the number of observation s.t. $y=1$ going to the left child, the number of observation s.t. $y=0$ going to the left child and similarly for the right child.
We now present the main loss functions that can be used which are adapted from standard DT losses.
\paragraph{Information Gain (ID3\cite{de1991distance},C4.5\cite{quinlan2014c4}C5.0\cite{im2005change})}
The information gain represents the amount by which the entropy of the class labels changes w.r.t the splitting of the dataset. It has to be maximized which happens when one minimizes the weighted sum of the local entropies, which in turn requires the class distribution of each region to converge to a Dirac. It is defined as
\begin{align}
IG(x,y;\Phi) = &E(P,N)-\frac{n_{left}}{P+N}*E(P_{left},H_{left})\nonumber \\
&-\frac{n_{right}}{P+N}*E(P_{right},N_{right}).
\end{align}

\paragraph{Gini Impurity (CART\cite{lewis2000introduction})}
The Gini impurity is more statistically rooted. It has to be minimized and attains the global minimum at $0$ happening when each region encodes only one class. It is defined for each region as
\begin{equation}
G(x,y)=1-\sum_kf_k^2,
\end{equation}
where $f_k$ is the proportion of observation of class $k$. It symbolizes the expected classification error incurred if a class label was drawn following the class label distribution of the leaf.
In fact, we have that
\begin{align*}
P(\hat{y}_i\not = y_i)=&\sum_kP(\hat{y}_i=k)P(y_i\not = k)\\
=&\sum_kf_k(1-f_k)\\
=&1-\sum_kf_k^2.
\end{align*}
The loss function per node is thus the weighted Gini impurity for each of the children defined as
\begin{align} 
i_{left}  = 1 - (\frac{P_{left}}{n_{left}})^2- (\frac{N_{left}}{n_{left}})^2,\\
i_{right}  = 1 - (\frac{P_{right}}{n_{right}})^2- (\frac{N_{right}}{n_{right}})^2,\\
G(x,y,\Phi)=\frac{n_{left}}{N+P}i_{left}+\frac{n_{right}}{N+P}i_{right},
\end{align}
with $N_k$ the number of observation of class $k$.

\paragraph{Variance Reduction (CART \cite{breiman1984classification})}
Finally, to tackle regression problems, another measure has been derived, the variance reduction. In this case, one aims to find the best split so that the intra-region variance is minimal. It is defined as
\begin{equation}
\frac{\sum_{i=1}^N\Phi(x_i)(y_i-\tilde{y}_1)^2}{\sum_{i=1}^N\Phi(x_i)}+\frac{\sum_{i=1}^N(1-\Phi(x_i))(y_i-\tilde{y}_2)^2}{\sum_{i=1}^N(1-\Phi(x_i))},
\end{equation}
with 
\begin{align}
\tilde{y}_1=\frac{\sum_{i=1}^N\Phi(x_i)y_i}{\sum_{i=1}^N\Phi(x_i)},\\
\tilde{y}_2=\frac{\sum_{i=1}^N(1-\Phi(x_i))y_i}{\sum_{i=1}^N(1-\Phi(x_i))}.
\end{align}
Note that a weighted version of this variance reduction can be used w.r.t the probability of each region.
Learning a NDT is now straightforward and similar to learning a DT except that now for each node instead of searching heuristically or exhaustively for the splitting criteria, it is optimized through an iterative optimization procedure such as gradient descend. This already alleviates the drawbacks of non differentiability encountered in oblique trees where GA had to be used to find the optimal hyperplane. However, it is also possible to go further by not just optimizing each splitting node in a greedy manner but optimizing all the splitting nodes simultaneously. In fact, with the Neural Decision Tree framework, we are now able to optimize the overall cost function simultaneously on all the nodes. This will loose the sequential aspect of the tests. However, this means that even though there exists no analytical solution for the global loss as it was the case in the greedy framework, the likelihood of being stuck in a local optimum is smaller. In fact, a non optimal split at a given node does not degrade all the children performances.
Thus the global loss function corresponds to the loss function of each last split node weighted by the probability to reach it. This is defined explicitly for the Gini impurity as
\begin{align}
Gini(Tree,\textbf{x})=&\sum_{leaf}\left[\frac{\sum_i P(x_i \in leaf)}{P+N}\right]\left(\frac{N_{leaf}}{P_{leaf}+N_{leaf}}\right. \nonumber \\
&\left.-\left(\frac{N_{leaf}}{P_{leaf}+N_{leaf}}\right)^2\right),
\end{align}
and for the entropy as
\begin{align}
IG(Tree,\textbf{x})=&E(P,N) \nonumber \\
&-\sum_{leaf}\left[\frac{\sum_i P(x_i \in leaf)}{P+N}\right]E\left( P_{leaf},N_{leaf}\right),
\end{align}
where the quantities $\left[\frac{\sum_i P(x_i \in leaf)}{P+N}\right]$ denote the probability that a given point belongs to this leaf which is estimated on the training set.
As a result these loss functions correspond to the generalization of the node loss function for all the leaves weighted by the probability to go into each of the leaves.

\section{Hashing Neural Network}
\subsection{Motivation}

One can see that for the special case where $G(x)=x^Tw+b$ we can rewrite the NDT as a perceptron where the output neurons all have a sigmoid function. The result of the output which will be an ordered chain $010011...$ is simply equivalent to the path of the corresponding tree that would put $x$ to the corresponding leaf as shown in Fig. \ref{fig1}.
\begin{figure}
\begin{center}
\includegraphics[width=3.5in]{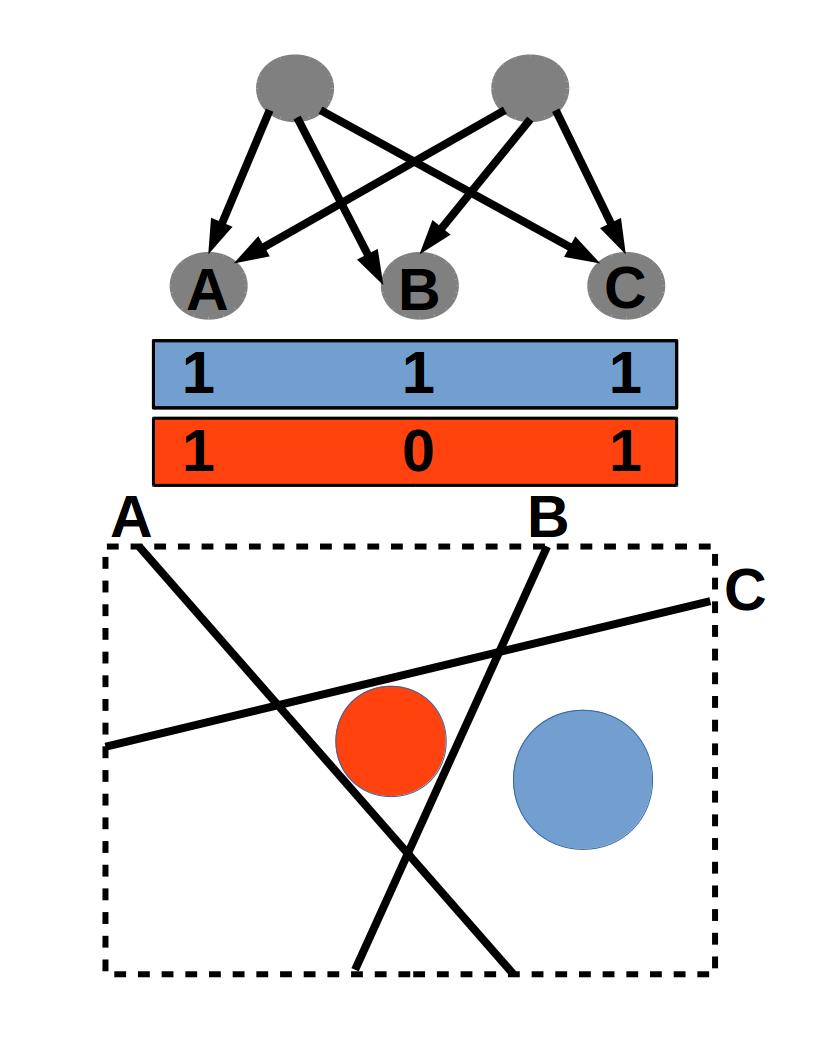}
\caption{Example of sub-region query}\label{fig1}
\end{center}
\end{figure}
This is the main motivation for rewriting the NDT as the HNN to then be able to add multiple layers and leverage deep learning framework combined with the NDT. This idea of combining the topology of DT with the learning capacities of ANN is not new. For example, combining ANN for latent space representation followed by DT is done in \cite{chandra2007combination}. Generating rules based on a trained ANN with DT is studied in \cite{fu1994rule,towell1993extracting,kamruzzaman2010rule,
craven1994using,craven1996extracting}. Using an ANN to filter a dataset prior to learning a DT is explored in \cite{krishnan1999extracting}. Finally and more recently, a reformulation of regression tree as a sparse ANN as been done in \cite{2016arXiv160407143B} in order to fine-tune the learned DT.
In our case, the HNN can be summarized simply as a generalized NDT in the sense that it will learn regions so that the class distribution within each of the region has the minimal uncertainty, ultimately with only one class per region but not necessarily one region per class as opposed to current architectures. In addition, we will see that thanks to the LSH framework, we are able to also perform HNN learning in a semi-supervised way or even unsupervised.
\subsection{The Hashing Layer}
In the HNN framework, the last layer now plays the role of an hashing function hence its hashing layer name. As a result, its output does not represent anymore $p(y|x)$ but $p(x\in L)$ where $L$ is any of the sub-region encoded by the network, and the output of the neurons correspond to a prediction of the path taken in a decision tree. From the sub-region membership, a prediction policy can be used based on the most present training data class present for example. This fact already highlights the ability to also make confidence prediction. In fact, if a region is ambiguous, different kind of predictions can be done based on the problem at hand. We no
\subsubsection{Supervised Case}
The number of output neurons $\#out$ must be of at least $\log_2(C)$ where $C$ is the number of classes in the task at hand. In fact, one needs a least $C$ different regions and it is clear that the number of different region that can be modeled by an hashing layer is $2^{\#out}$. For the case where the number of output neurons is greater, then the ANN has the flexibility to learn different sub-regions for each class. This is particularly interesting for example if the latent representation is somehow clustered yet the number of cluster is still greater than the number of classes and the cluster of same class are not necessarily neighbors.
In order to derive our loss function for multiclass problems we first need to impose the formulation of $y$ as a one-hot vector with a $1$ at the index of the class. From that we have the following quantities
\begin{align}
&out_n(x)=\Phi(x^Tw_n+b_n),\forall n\in\{1,...,\#out\}\\
&P=\{0,1\}^{\#out}
\end{align}
and finally we call $chain$ each element of $P$ which fully and uniquely identify each of the sub-regions encoded in the network.
We thus have 
\begin{equation}
p(x_i \in chain)=\prod_n out_n(x_i)^{1_{chain(n)=1}}(1-out_n(x_i))^{1_{chain(n)=0}}
\end{equation}
the average $y$ per sub-region is thus the weighted mean of all the $y$ w.r.t their membership probability, this will give the class distribution per $chain$ as 
\begin{equation}
y_{chain}=\sum_iy_ip(x_i\in chain),
\end{equation}
where we recall that $y_i$ is a one-hot vector and thus $y_{chain}$ is a vector of size $C$ which sums to $1$ and is nonnegative. Thus it is $p(y|chain)$.
Finally, we can now compute the measure of uncertainty $E$ for each $chain$ whether it is the entropy, the Gini impurity or any other differentiable function and write the final loss function as the weighted sum of these uncertainties weighted by the probability to reach each sub-regions as
\begin{equation}
loss=\sum_{x \in X}\sum_{chain \in P}p(chain)E(y_{chain}),\label{loss1}
\end{equation}
where as for the NDT the estimated probability to reach a sub-region is simply
\begin{equation}
p(chain)=\frac{\sum_{i=1}^Np(x_i \in chain)}{N}.
\end{equation}
Note that all the classification loss can be used with class weights in order to deal with unbalanced dataset which occur often in real world problems.
For regression problems we first define the local variance of the outputs as
\begin{equation}
V(leaf)=\frac{\sum_i (y_i-\tilde{y}_{chain})^2p(x_i \in chain)}{\sum_ip(x_i \in chain)},
\end{equation}
where
\begin{equation}
\tilde{y}_{chain}=\frac{\sum_i y_i p(x_i \in chain)}{\sum_i p(x_i \in chain)}.
\end{equation}
As a result we can rewrite the complete regression loss as
\begin{equation}
loss =\sum_{x \in X} \sum_{chain \in P}p(chain)V(leaf).
\end{equation}
Note that it is possible to not weight the local loss functions by the probability to reach the region.

Finally, note that this hashing layer can obviously be used after any already known neural network layer such as densely connected \cite{gardner1998artificial} or convolutional \cite{krizhevsky2012imagenet} layers or even recurrent based layers \cite{gers2000learning} making HNN acting on deep and latent representation rather than the input space. The training is thus done through back-propagation with the presented loss.
\subsubsection{Semi-Supervised Learning}
One extension not developped in the paper concerns the semi-supervised case for which the standard discriminative loss is used for labeled examples s.a. the Information Gain of the Entropy and the unsupervised loss is used for the unlabbeled examples as well as the labeled examples. The unlabeled loss would tipycally be the intra region variance which is similar to the loss of k-NN algorithm or GMM with identity covariance matrix. This result in aggregating into same region parts of the space with high density while constraining that two different labels never occur inside one region. Extensions of this will be presented as well as validation results in order to validate this hybrid loss between Decision Trees and k-NN.
\subsubsection{Unsupervised Case}
In the last section a loss function was derived for the case where we have access to all the labels $y_i$ of the training inputs $x_i$. 
The semi-supervised framework is basically a deep-autoencoder on which a hashing layer is connected to the latent space, namely the middle layer. As a result, the reconstruction loss is applied to all inputs and the standard hashing layer loss is used when labels are available.
For unsupervised however, the deep autoencoder is trained again coupled with an hashing layer but this time the hashing layer is unsupervised.
We now derive analytically the loss function of the hashing layer in the unsupervised case. There are two ways to do it, it can either be random to become a known LSH function such as MinHash or one can try to find clusters so that the intra-cluster variance is minimal similarly to a k-NN \cite{fukunaga1975branch} approach.
The variance per region is defined as
\begin{equation}
V_{chain}(x)=\frac{\sum_i(x_i-\tilde{x}_{chain})^2p(x_i\in chain)}{\sum_i p(x_i\in chain)},
\end{equation}
where the local sub-region means is defined as
\begin{equation}
\tilde{x}_{chain}=\frac{\sum_i p(x_i \in chain)x_i}{\sum_i p(x_i \in chain)}.
\end{equation}
As a result, the overall loss is simply
\begin{equation}
loss=\sum_{x \in X}\sum_{chain \in P}p(chain)V(x).
\end{equation}
If we denote by $G$ the last layer of the neural network before the hashing layer, the final unsupervised loss thus becomes
\begin{equation}
loss = \sum_{x \in X}||G^{-1}(G(x))-x||^2+\sum_{chain \in P}p(chain)V(G(x)),
\end{equation}
and for the semi-supervised case
\begin{align}
loss = &\sum_{x \in X}||G^{-1}(G(x))-x||^2\nonumber \\
&+\sum_{chain \in P}p(chain)E(G(x))1_{\text{$x$ has a label}}.
\end{align}
Concerning the random strategy, similarly to Extremely Randomized Trees, it is possible to adopt an unlearn approach for which the hyperplanes $w_b$ are drawn according to a Normal distribution. This way, we follow the LSH framework for which we have
\begin{equation}
p(out_n(x_i)\not =out_n(x_j)) \propto d(x_i,x_j),
\end{equation}
where $d$ is a distance measure.

\subsection{Training}
\subsubsection{Iterative Optimization Schemes}

Since there is no analytical form to find the optimal weights and bias inside a deep neural network, one has to use iterative optimization methods. Two of the main possibilities are Genetic Algorithms \cite{davis1991handbook} and Gradient based methods. We focus here on the advances for gradient based methods as it is the most popular optimization technique nowadays.
Put simply, the update for each free parameter $W$, the update rule is
\begin{equation}
W = W -\alpha \nabla W + \beta f(W),
\end{equation}
where $\alpha$ is the learning rate and $\beta$ a regularization parameter applied on some extra function $f$. A common technique to find the best learning rate and regularizer is cross-validation but new techniques have been developed allowing an adaptive learning rate and momentum which are changed during training\cite{yu2006adaptive,yu2002backpropagation,
hamid2011accelerating,nawi2011enhancing}.
Whatever activation function is used, one can also add new parameters to the input in order to scale the input as presented in \cite{he2015delving}.
Finally, many tricks are studies for better back-propagation in \cite{lecun2012efficient} and a deep study of the behavior of the weights during learning is carried out in \cite{lecun1991second}.
We now derive the explicit gradient for the case where the loss $E$ is the Gini impurity.
It is clear that 
\begin{align}
\frac{d }{d W}loss = &\sum_n\frac{d loss}{d out_n}\frac{d out_n}{d W} \nonumber \\
=&\sum_{chain \in P}\sum_n \frac{d p(chain)E(y_{chain})}{d out_n}\frac{d out_n}{d W} 
\end{align}
with $\frac{d p(chain)E(y_chain)}{d out_n}$ a scalar and $\frac{d out_n}{d W}$ a matrix. We now derive explicitly the derivative:
\begin{align}
\frac{d out(x_i)}{dW}=
\left(
\begin{matrix}
(-1)^{1_{chain(1)=0}}out_1(x_i)(1-out_1(x_i))x_i^T\\
(-1)^{1_{chain(2)=0}}out_2(x_i)(1-out_2(x_i))x_i^T\\
\vdots
\end{matrix}\right),
\end{align}
which is of size $(\#out,\#in)$ and is basically on each row the input $x_i$ which might be the output of another upper layer weighted by the activation, it is similar to a sigmoid based layer except for the indicator function which helps to determine the sign.
Now if we now denote the true output by
\begin{equation}
\sigma^{chain(n)}_n(x_i)=out_n(x_i)^{1_{chain(n)=1}}(1-out_n(x_i))^{1_{chain(n)=0}},
\end{equation}
which includes the indicator function for clarity, we have
\begin{align}
\frac{\partial }{\partial out_n} &p(chain)=\frac{\partial }{\partial out_n} \frac{\sum_i \prod_n \sigma_n(x_i)}{N} \nonumber\\
&=\frac{\sum_i\sum_k(-1)^{1_{chain(k)=0}}\prod_{n\not = k} \sigma^{chain(n)}_n(x_i)}{N}.
\end{align}
Note that the derivative of the probability w.r.t the output is thus simply a sum of the products of the other output neurons where the neuron considered in the derivative determines the sign applied. In short, we see that this changes linearly as one fixes all the neurons but the one considered for variations, which is natural.
We now derive the final needed derivate:
\begin{equation}
    \begin{aligned}
&\frac{\partial }{\partial out_n} E(y_{chain})=\frac{\partial }{\partial out_n}(1-\sum_k f_k^2) \nonumber\\
=&-\sum_k \frac{\partial }{\partial out_n} \left(\frac{\sum_iy_{i,k}p(x_i \in chain)}{\sum_j \sum_i y_{i,j}p(x_i \in chain)}\right)^2 \nonumber \\
=&-\sum_k \frac{\partial }{\partial out_n} \left(\frac{\sum_iy_{i,k}\prod_n \sigma^{chain(n)}_n(x_i)}{\sum_j \sum_i y_{i,j}\prod_n \sigma^{chain(n)}_n(x_i)}\right)^2 \nonumber \\
=&-\sum_k 2f_k \left(\frac{(\sum_iy_{i,k}\sum_l(-1)^{1_{chain(l)=0}}\prod_{n\not = l} \sigma^{chain(n)}_n(x_i))}{(\sum_j \sum_i y_{i,j}\prod_n \sigma^{chain(n)}_n(x_i))^2}\right. \nonumber \\
&\times \left. (\sum_j \sum_i y_{i,j}\prod_n \sigma^{chain(n)}_n(x_i)) \right) \nonumber \\
&+\sum_k 2f_k \left(\frac{(\sum_iy_{i,k}\prod_n \sigma^{chain(n)}_n(x_i))}{(\sum_j \sum_i y_{i,j}\prod_n \sigma^{chain(n)}_n(x_i))^2}\right. \nonumber \\
&\times \left. (\sum_j \sum_i y_{i,j}
\sum_l(-1)^1_{chain(l)=0}\prod_{n\not = l} \sigma^{chain(n)}_n(x_i)) \right) 
\end{aligned}
\end{equation} 
Finally, note that performing stochastic gradient descent in an option when dealing with large training set. It has also been shown to improve the convergence rate.

\subsubsection{Regularization Techniques}
Initially, regularization was done by adding to the standard cost function a regularization term, typically the L1 or L2 norm of the weights. This imposed the learned parameters to be sparse. From that new kind of regularization techniques have been developed such as dropout \cite{hinton2012improving}, \cite{srivastava2014dropout}. In dropout, each neuron has a nonzero probability to be deactivated (simply output $0$) forcing the weights to avoid co-adaptation. This probabilistic deactivation can be transformed into adding some Gaussian noise to each of the neuron outputs which in this case force the weights to be robust to noise. 
The motivation here is not necessarily to avoid overfitting, in fact, if using ensemble methods, over fitting is actually better to then leverage variance reduction from averaging \cite{krogh1996learning}.
However another type of regularization can be used on the distribution of the data across the regions. One such type might be
\begin{equation}
\sum_{chain \in P} ||p(chain)-\frac{1}{|P|}||^2
\end{equation}
so that regions become more equally likely.
\subsection{Toy Dataset}
We now present the application of the HNN on two simple toy datasets for binary classification, the two moons and circle dataset. Each one presents nonlinearly separable data points yet the boundary decision can very effectively be represented by a small union of linear plans. The main result is that the number of parameters needed with the HNN is smaller than when using an ANN. For all the below examples, the ANN is made of 3 layers with topology $2:3:1$ which is the smallest network able to tackle these two problems. As we will see, the HNN only requires one layer and the number of needed parameters for similar decision boundary goes for from $13$ for ANN to $9$ for HNN. This simple example shows that the reformulation of the hashing layer helps to avoid overfitting in general. In fact, overfitting is not necessarily learning the training set but it is also using a more flexible model that needs be \cite{hawkins2004problem}. Since with the HNN we are able to obtain the same decision boundaries yet with less parameters, it means that the ANN architecture was somehow sub-optimal.
\subsubsection{Two-moon Dataset}
\begin{figure}[t!]
\begin{center}
\includegraphics[width=3.5in]{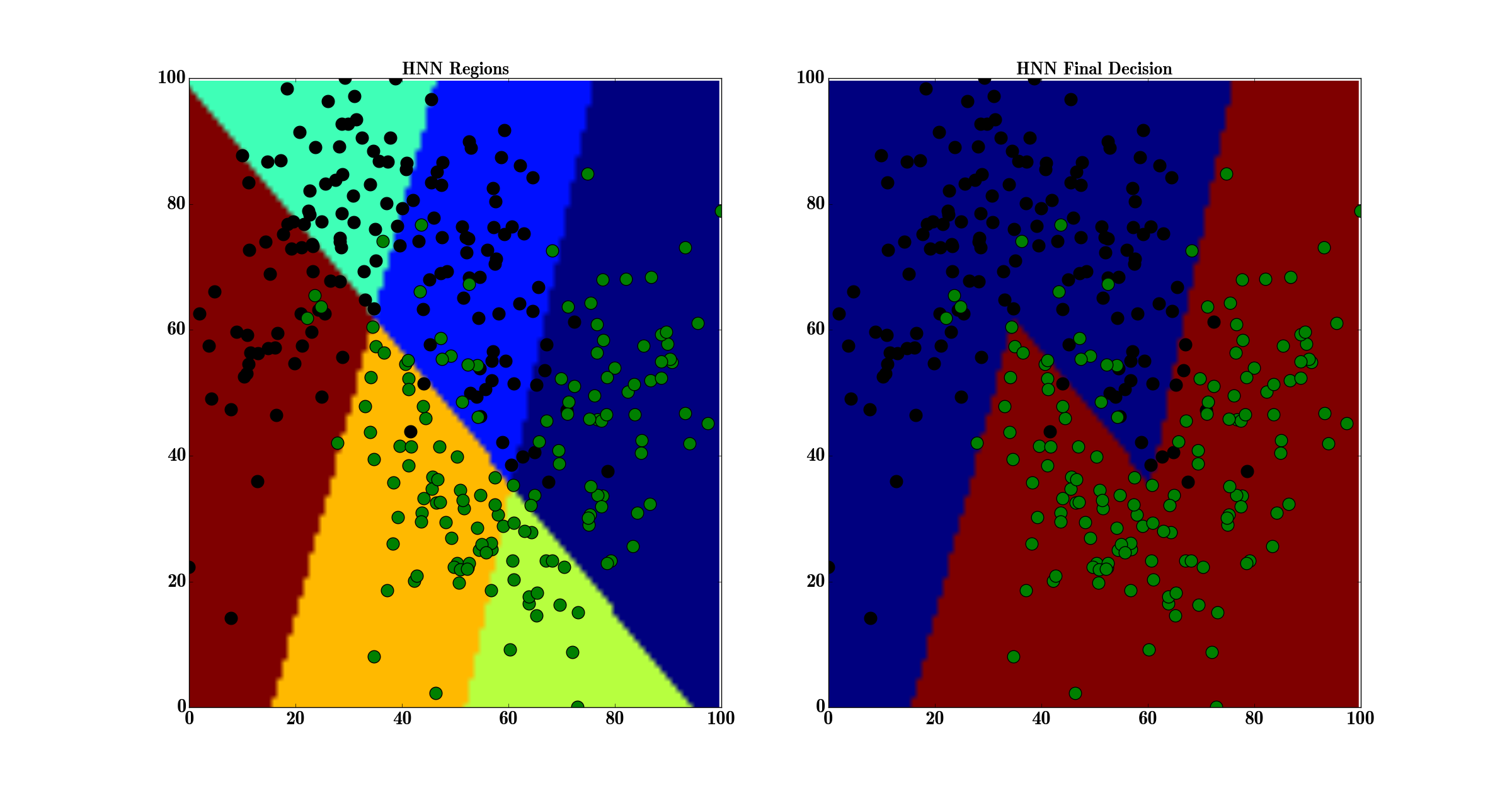}
\caption{Noisy Two-Moons Dataset, learned regions and final binary classification regions for HNN.}\label{fig11}
\end{center}
\end{figure}

\begin{figure}[t!]
\begin{center}
\includegraphics[width=3.5in]{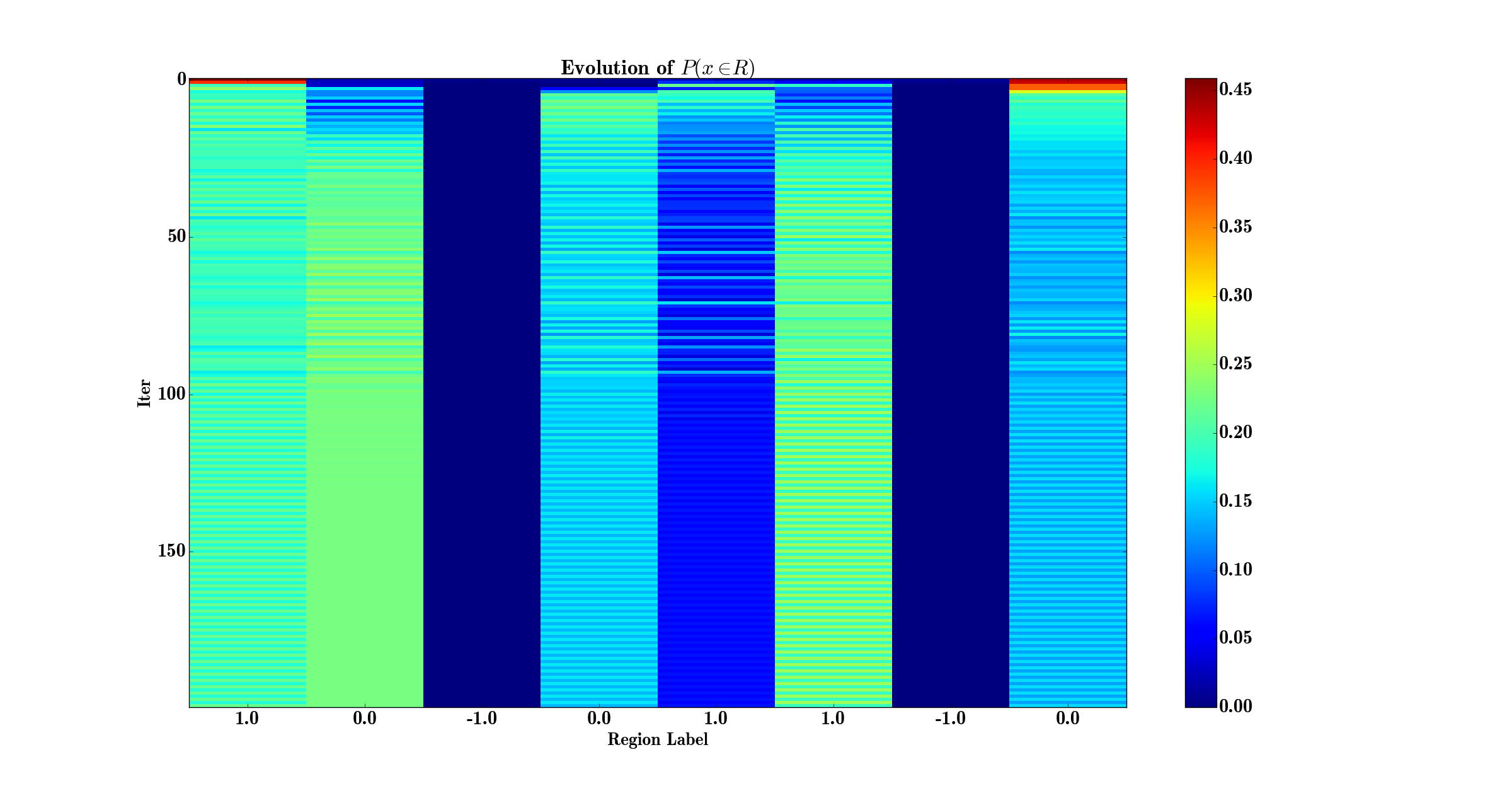}
\caption{Evolution of $p(chain)$ over the $8$ regions learned during training.}\label{fig22}
\end{center}
\end{figure}

\begin{figure}[t!]
\begin{center}
\includegraphics[width=3.5in]{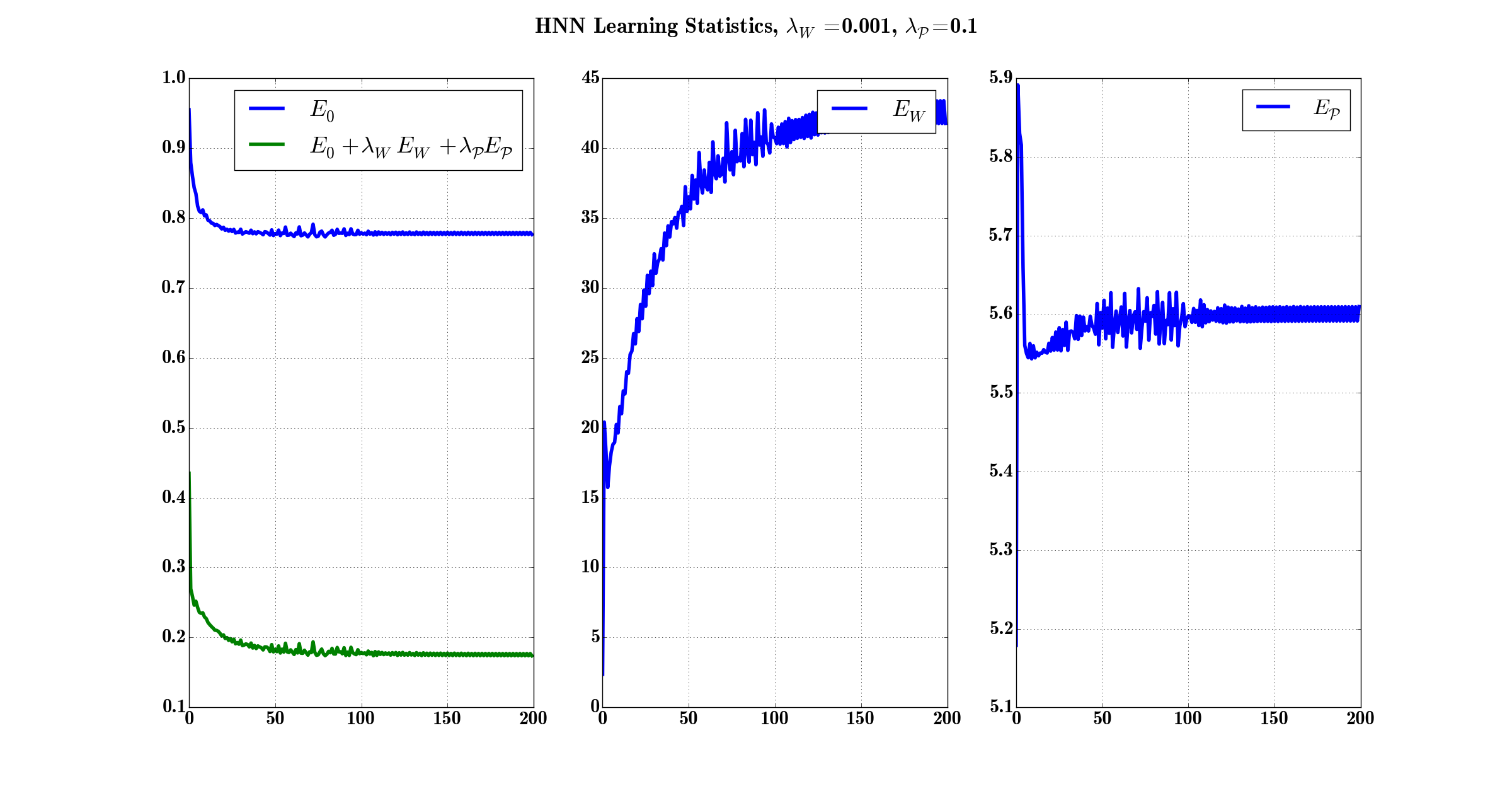}
\caption{Evolution of the errors during training. On the left is the pure error as defined in \ref{loss1} and with addition of the regularization terms. On the middle is the norm of the weight and on the right the distance with respect to a uniform distribution of the points in each region.}\label{fig33}
\end{center}
\end{figure}

\begin{figure}[t!]
\begin{center}
\includegraphics[width=3.5in]{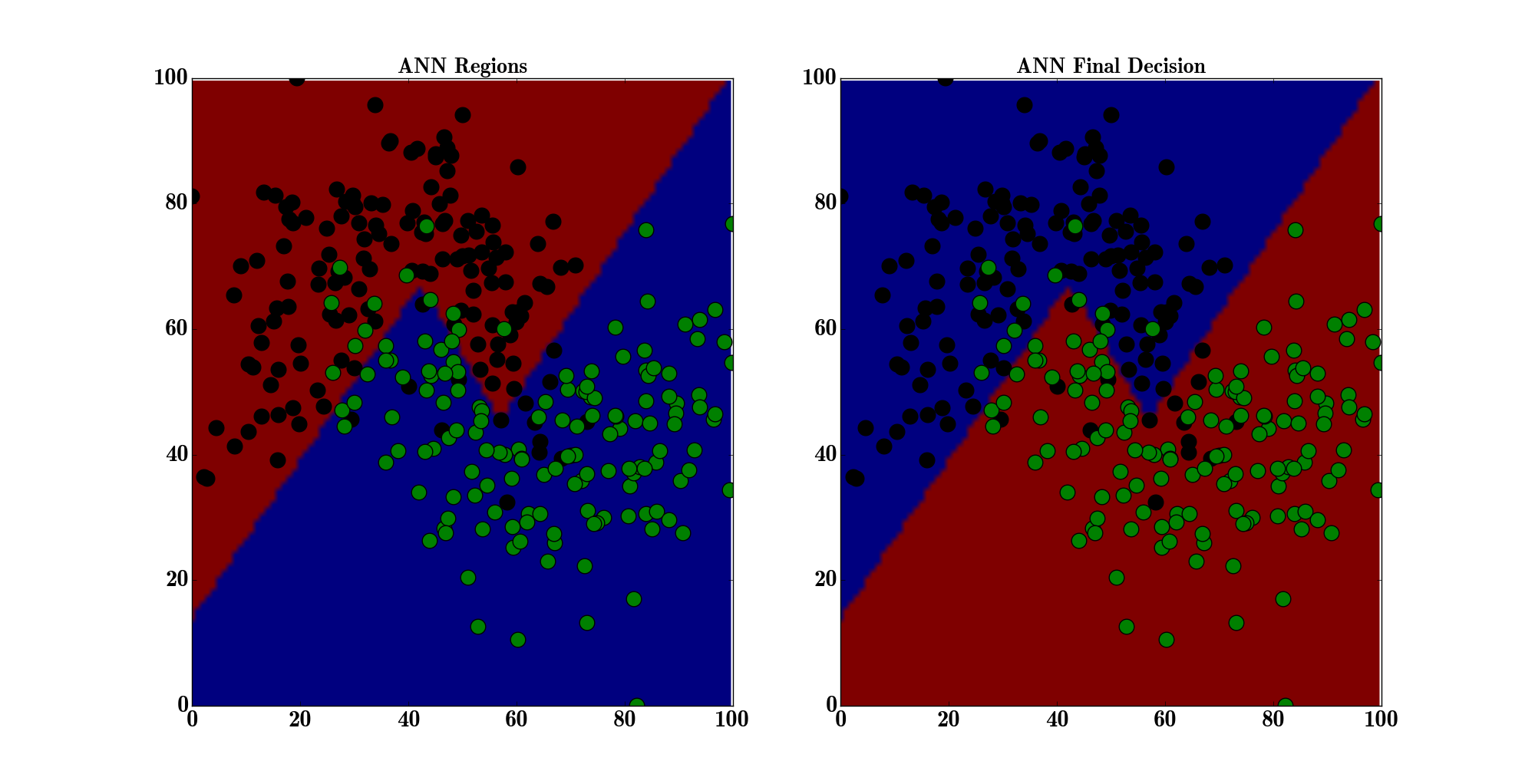}
\caption{Noisy Two-Moons Dataset, learned regions and final binary classification regions for HNN.}\label{fig111}
\end{center}
\end{figure}

\begin{figure}[t!]
\begin{center}
\includegraphics[width=3.5in]{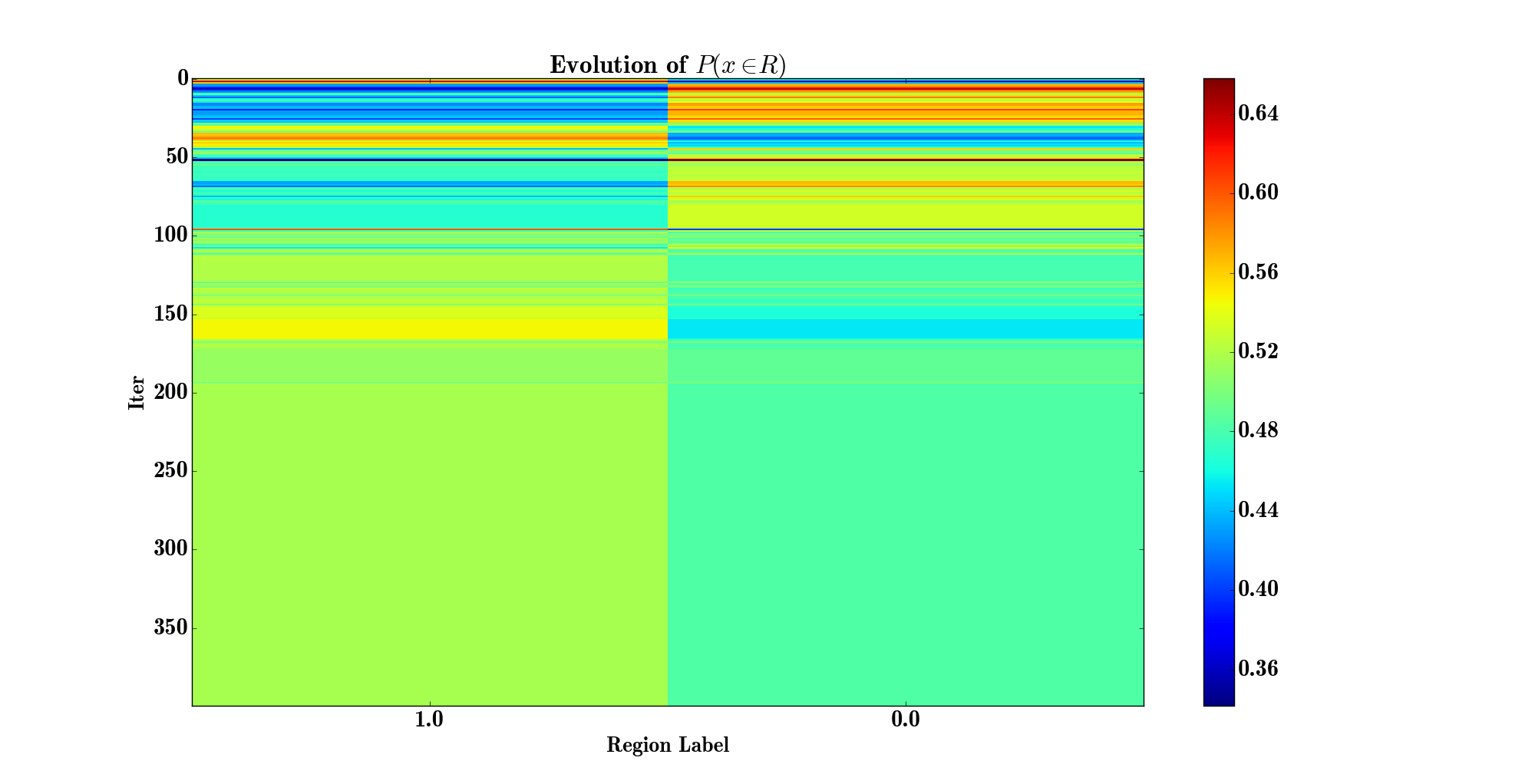}
\caption{Evolution of $p(y=0)$ and $p(y=1)$ during training of the MLP.}\label{fig222}
\end{center}
\end{figure}

\begin{figure}[t!]
\begin{center}
\includegraphics[width=3.5in]{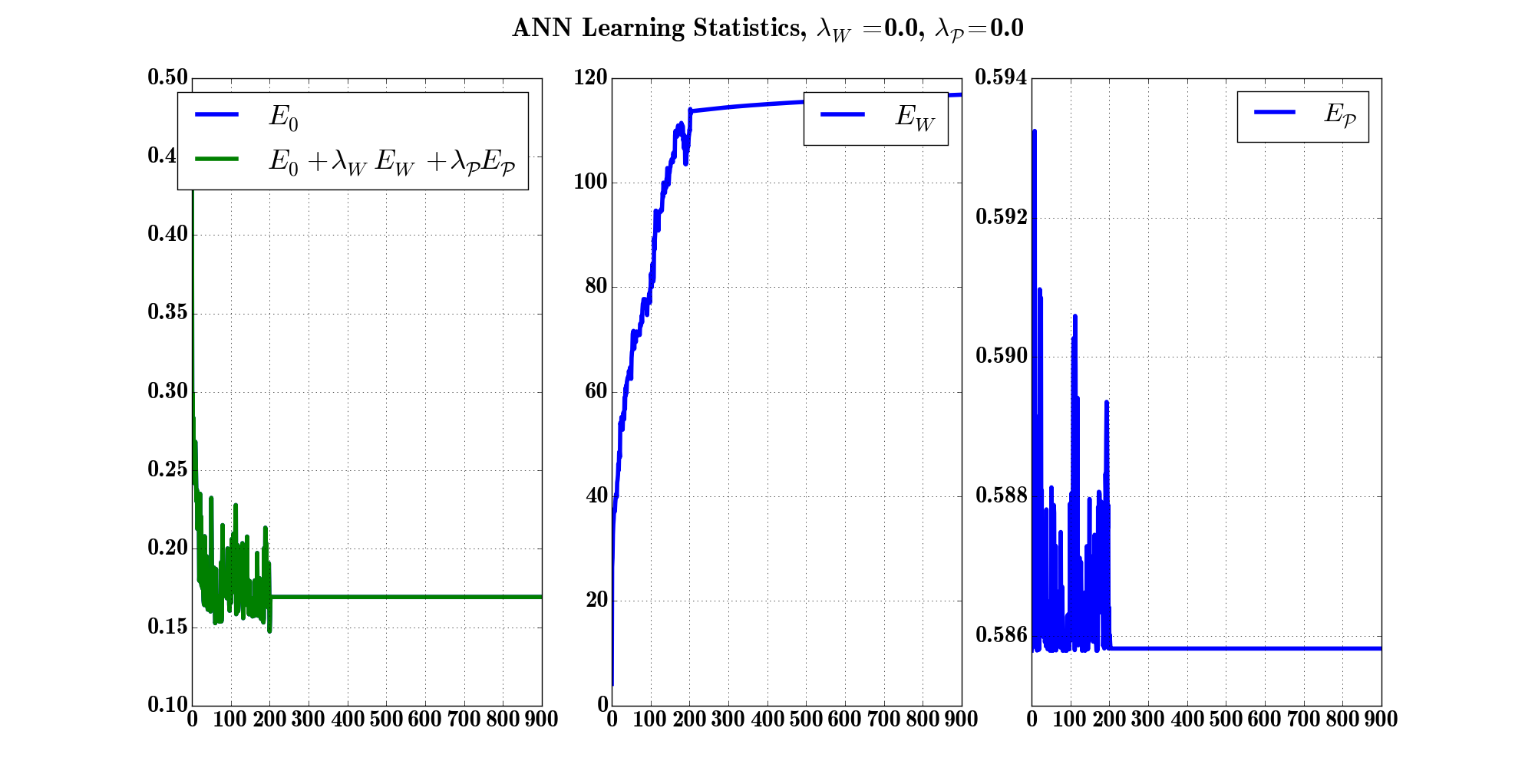}
\caption{Evolution of the errors during training for the ANN.}\label{fig333}
\end{center}
\end{figure}

The two-moon dataset is a typical example of nonlinear binary classification. It can be solved easily with kernel based methods or nonlinear classifiers in general. As we will show, even though the used HNN has only one layer, by the way it combines the learned hyper-plans it is possible to learn nonlinear decision boundaries. In Fig. \ref{fig11} one can see the HNN with $\#out=3$ after training. The boundary decision is similar in shape with the one learned form a $2:3:1$ MLP shown in Fig. \ref{fig111}. In fact, it is made up of $3$ combined hyper-plans.
One can see in Fig. \ref{fig22} the evolution of the probability to reach each of the $8$ regions during training as opposed to the $2$ regions for the case of the MLP presented in Fig. \ref{fig222}.
We also present in Fig. \ref{fig33} the evolution of the error and regularizations during training for the HNN. As one can see, regions starting with almost no points can recover and become preponderant whereas useful regions can be disregarded at any point in the training. Similarly we have for Fig. \ref{fig333} the case with the ANN. It is interesting to see that the convergence rate is also faster with the HNN. In fact it converges in $50$ iterations whereas ANN converges in around $150$ iterations.
\subsubsection{Two-circle Dataset}
The two-circle dataset is another simple yet meaningful dataset. It presents two circles with same center with different radius, as a consequence one is inside the other and the binary classification task is to discriminate between the two. It is quite straightforward to see that a hand-craft change of variable $(x,y)\rightarrow(r,\theta)$ can make this problem linearly separable yet we will see how HNN and ANN solve this discrimination problem.
\begin{figure}[t!]
\begin{center}
\includegraphics[width=3.5in]{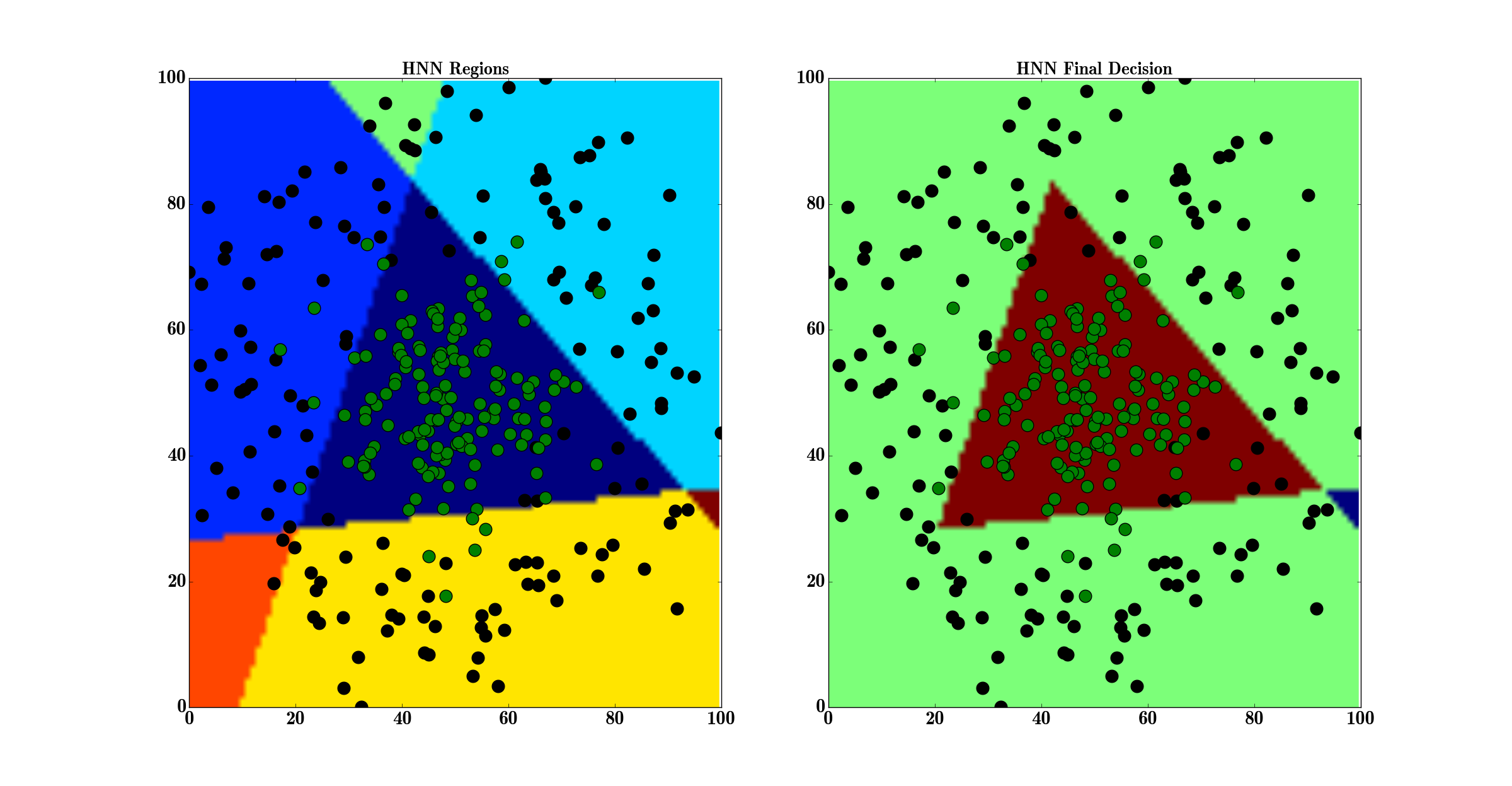}
\caption{Regions and decision boundary for the HNN in the two-circle dataset.}\label{fig1}
\end{center}
\end{figure}
We present in Fig. \ref{fig1} the result of the HNN with $3$ neurons. Again, the decision boundary is also presented for the case of an ANN with topology $2:3:1$ in Fig. \ref{fig1111}.
\begin{figure}[t!]
\begin{center}
\includegraphics[width=3.5in]{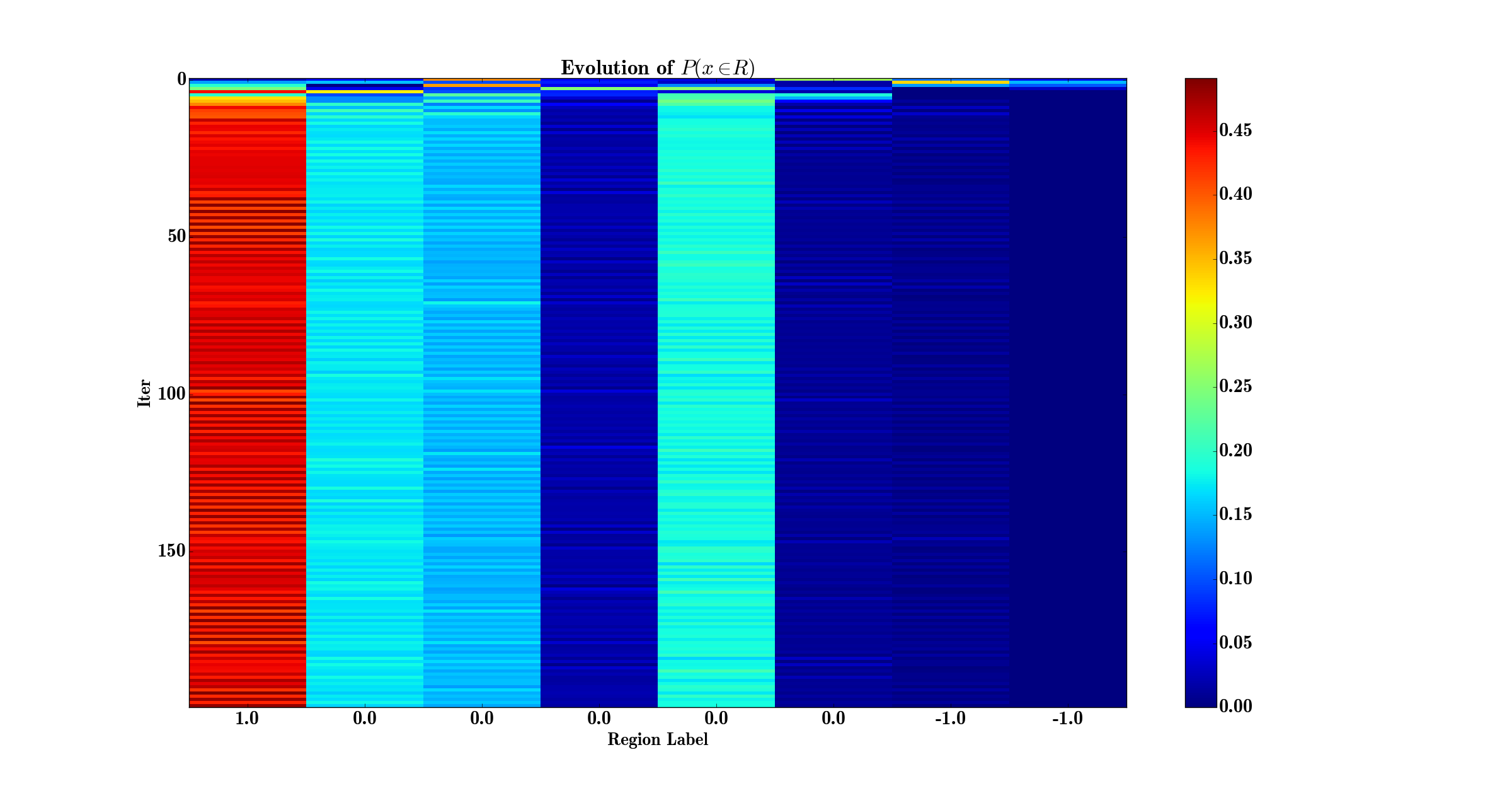}
\caption{Evolution of $p(chain)$ during training for the $8$ possible regions of the HNN on the two-circle dataset.}\label{fig2}
\end{center}
\end{figure}
We also present in Fig. \ref{fig2} for the HNN and Fig. \ref{fig2222} for the ANN the evolution of the probability to reach the sub-regions during training. As can be seen in Fig. \ref{fig3} and \ref{fig3333} the convergence rate is significantly faster for the HNN. In fact the convergence is done in about $20$ iteration whereas the neural network needs a bit more than $300$ iterations.
\begin{figure}[t!]
\begin{center}
\includegraphics[width=3.5in]{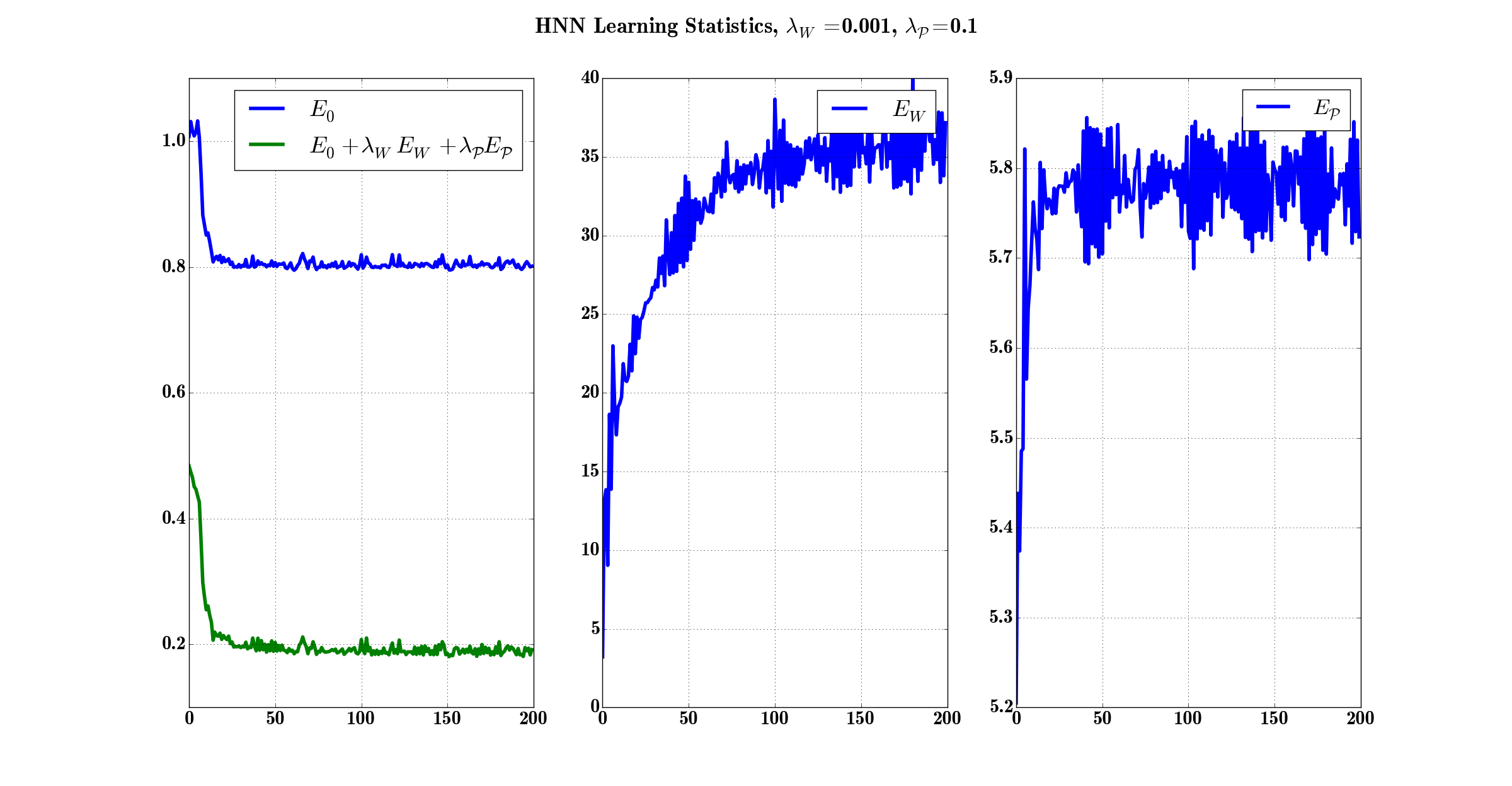}
\caption{Evolution of the error and regularization for the HNN with $3$ neurons.}\label{fig3}
\end{center}
\end{figure}
\begin{figure}[t!]
\begin{center}
\includegraphics[width=3.5in]{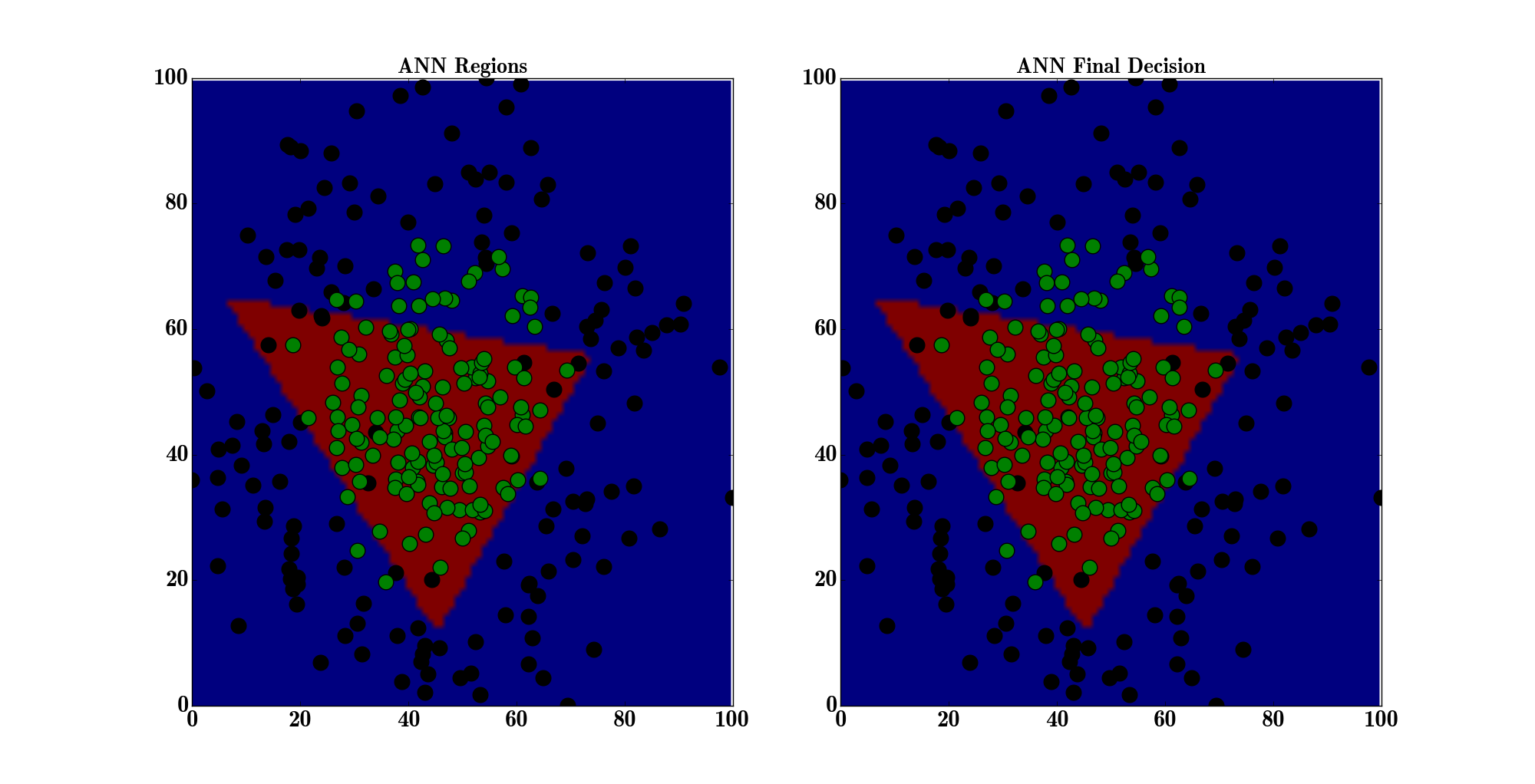}
\caption{ANN Decision Boundary for the two-circle dataset}\label{fig1111}
\end{center}
\end{figure}
\begin{figure}[t!]
\begin{center}
\includegraphics[width=3.5in]{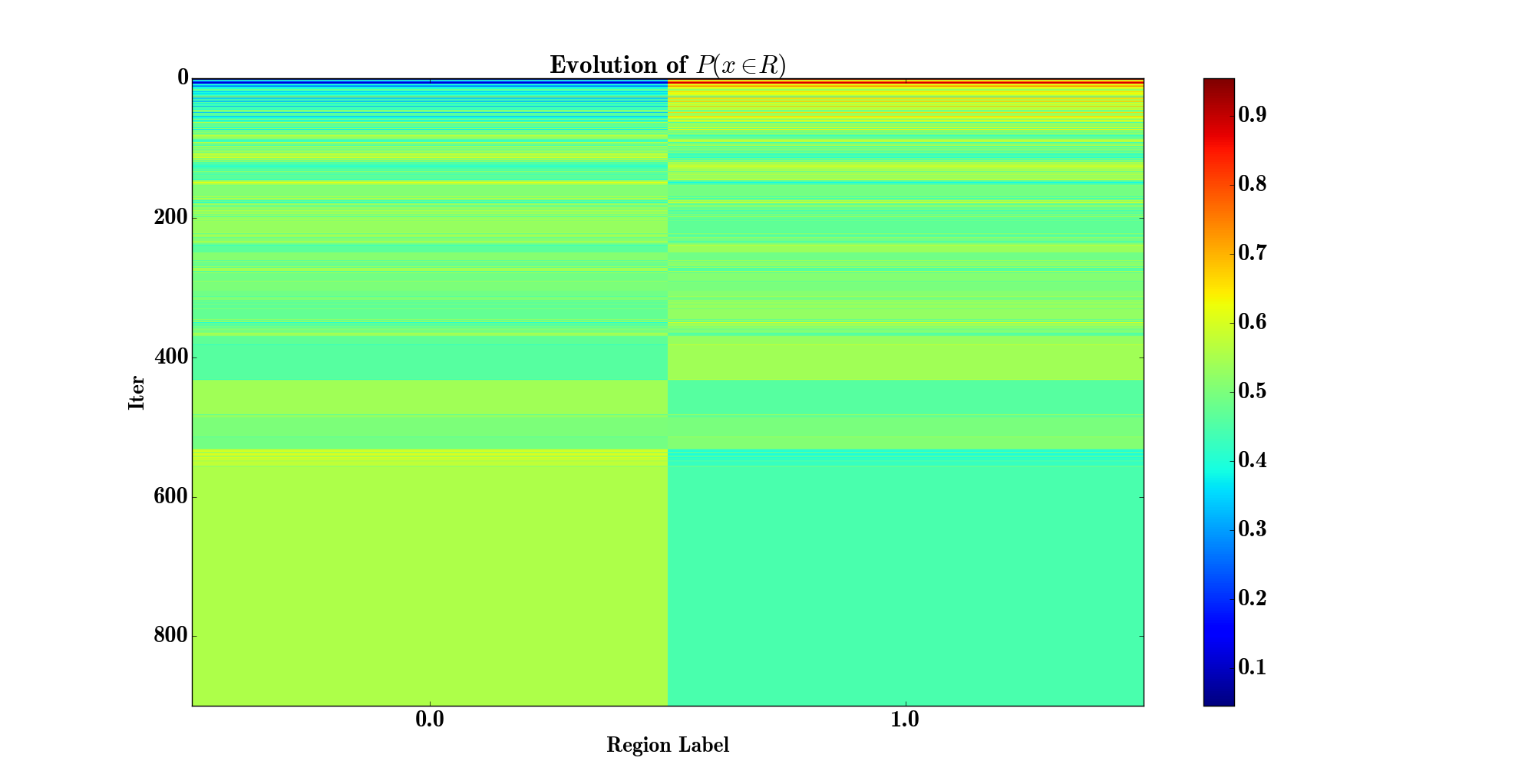}
\caption{ANN $p(chain)$ for the two classes}\label{fig2222}
\end{center}
\end{figure}
\begin{figure}[t!]
\begin{center}
\includegraphics[width=3.5in]{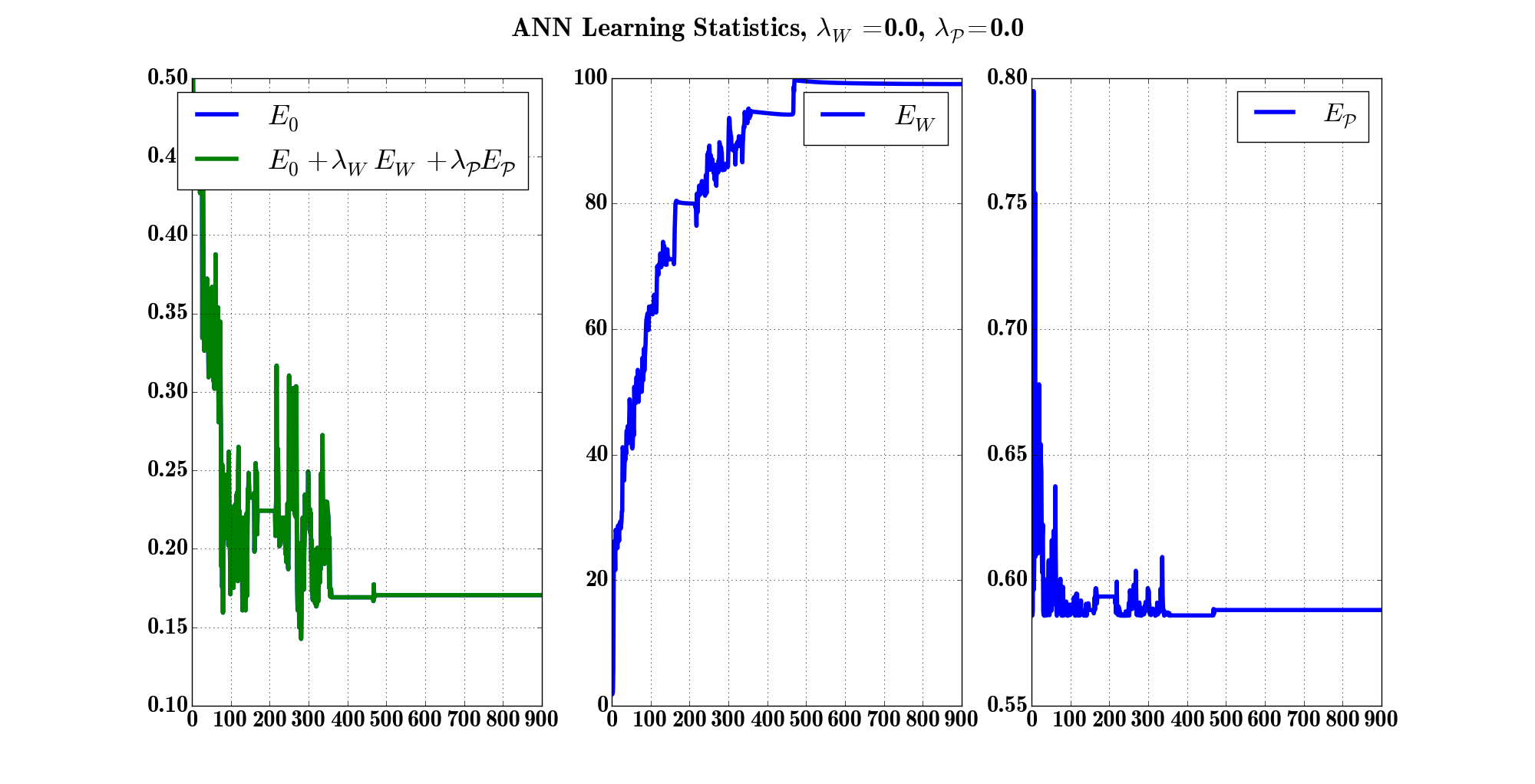}
\caption{ANN training statistics including the error and the regularizations}\label{fig3333}
\end{center}
\end{figure}
Finally, we present the case where we used $4$ neurons for the HNN on the two-circle dataset. Note that this puts the number of parameters to $12$ which is still less than the number of parameters used in the ANN $13$. Yet if one wanted to have the same modeling ability with a MLP, the minimum topology would be $2:4:1$ which contains $17$ parameters. In fact, it grows exponentially w.r.t the number of hidden neurons whereas for the HNN it grows linearly since the modeling capacity grows exponentially. We present in Fig. \ref{fig0}, Fig. \ref{fig00} and Fig. \ref{fig000} the results for the $4$ neurons HNN.
\begin{figure}[t!]
\begin{center}
\includegraphics[width=3.5in]{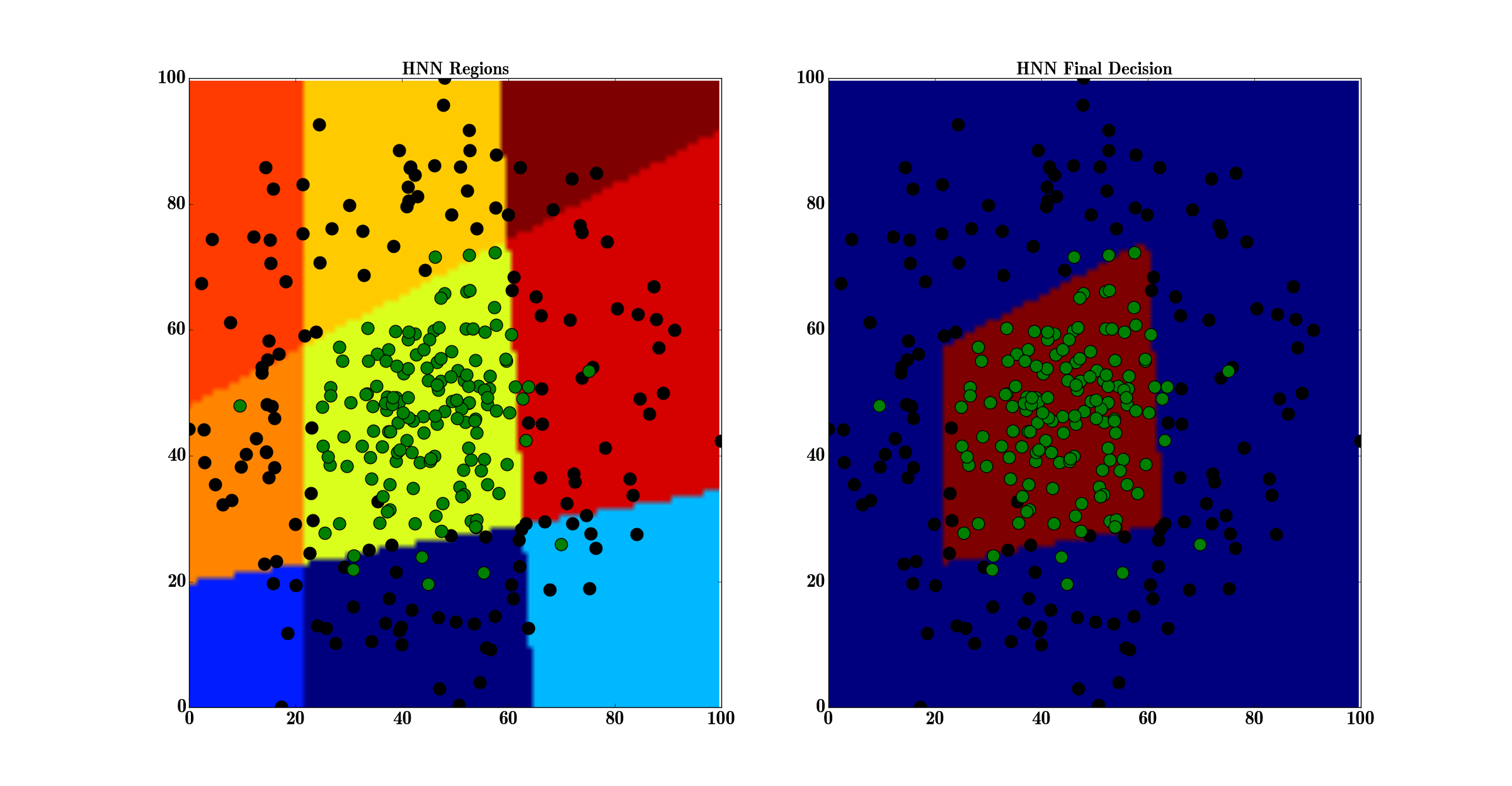}
\caption{$4$ neurons HNN}\label{fig0}
\end{center}
\end{figure}
\begin{figure}[t!]
\begin{center}
\includegraphics[width=3.5in]{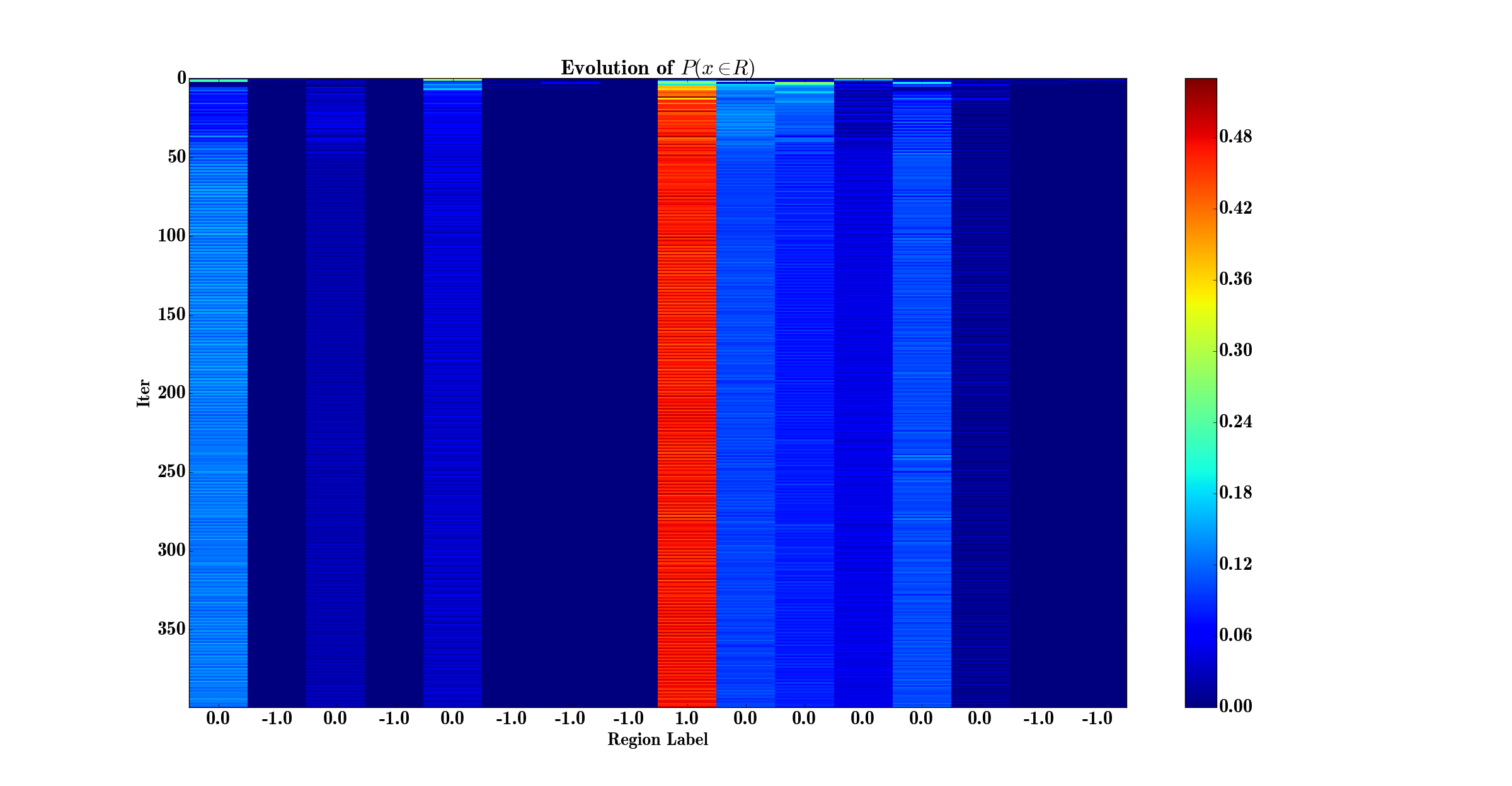}
\caption{$4$ neurons HNN, evolution of $p(chain)$.}\label{fig00}
\end{center}
\end{figure}
\begin{figure}[t!]
\begin{center}
\includegraphics[width=3.5in]{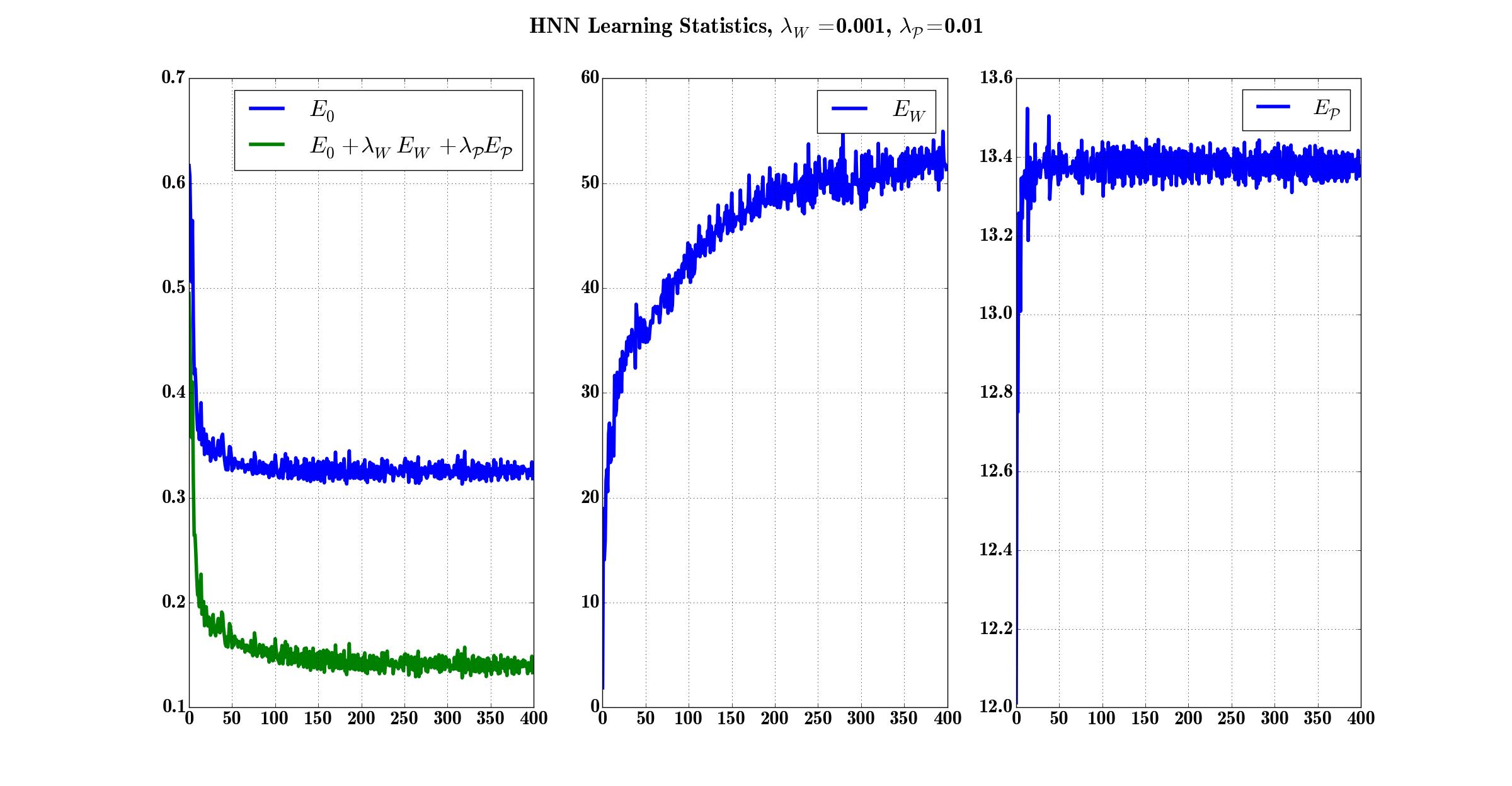}
\caption{$4$ neurons HNN, evolution of the error and regularization errors during training.}\label{fig000}
\end{center}
\end{figure}

\section{Future Work}
This paper introduces a new way to improve neural networks through the analysis of the output activations and the loss function. In fact, it has been demonstrated for example in \cite{tang2013deep} a SVM type loss is used instead of the standard cross-entropy. As a result the generalization loss has been diminished. Yet the important point was that the cross-entropy of the new trained network was far from optimal or even close to what one could consider as satisfactory. This suggests that different loss functions do not just affect the learning but also the final network. As a result, one part of the future work is to study the impact of the loss function on different training set with fixed topologies. This includes the training with standard neural networks and the HNN for the case where $\#out=\log_2(n)$. This could lead to a new framework aiming at learning the loss function online.

With the analysis of trees one natural extension of the HNN is boosting or its simpler form, bagging. Doing this ensemble methods with complete ANN might be difficult due to high complexity to already train one network. As a result, some techniques such as dropout have been used and analyzed as a weak way to perform model averaging or bagging. A solution in our case would be to perform bagging or boosting of only the hashing layer part, namely the last layer of the HNN. This way the workers live on the same latent space but act on different pieces of it and their combination is used for improving the latent space representation.
Another approach would be to see model averaging from a bayesian point of view \cite{penny2006bayesian}. The model evidence is given by 
\begin{equation}
p(y|m)=\int p(y|\theta,m)p(\theta|m)d\theta,
\end{equation}
and the Bayesian Model Selection (BMS) is
\begin{equation}
m_{MP}=\argmax_{m \in M}p(m|y).
\end{equation}
When dealing with one model $m$, the inference $p(\theta|y,m)$ is dependent w.r.t to the chosen model. In order to take into account the uncertainty in the choice of the model, model averaging can be used. Model Averaging (BMA) formulates the distribution
\begin{equation}
p(\theta|y)=\sum_mp(\theta|y,m)p(m|y),
\end{equation} 
look for a review in \cite{hoeting1999bayesian}.
This whole framework can be used with a nonparametric PGM with a new define probability for a neural network, and $p(x|ANN)$ as the reconstruction error and $p(ANN)$ based on the model complexity for example or the topology of the connections.

Finally, an important aspect resides in the correlation between different $out_n(x_i)$ and $out_m(x_i)$ for different inputs. In fact, they have to be not correlated otherwise it means that the hyperplanes are scaled versions of each others.

\section{Conclusion}

In this paper we presented an extension of DTs and ANNs to provide a unified framework allowing to take the best of both approaches. The ability to hash a dataset into an exponentially large number of regions which is the strenfgth of DT coupled with the ability to learn any boundary decision for those region coming from the ability of ANN to model arbritrary functions. We leverage the differentiable of our approach to derive a global loss function to train all the nodes simultaneously with respect to the resulting leaves entropy showing robustness to poor local optimum DTs can fall in. The differentiability of the model allows easy integreation in many machine learning pipeline allowing the extension of CNNs to robust semi-supervised clustering for example. In addition, the ability to learn arbitrary union of regions of the space to perform a per class clustering reduces the necessary condition of having fully linearized the dataset. This shall reduce the required depth of today's deep architectures. Finally, this network has now the capacity to be information theoretically optimal as the minimum required number of neurons in supervised problems is $\log_2(C)$ as opposed to $C$ for usual soft-max layers where $C$ is the number of classes. Finally, the possibility to apply this framework in supervised as well as unsupervised settings might lead to interesting behavior through the clustering property of the latent representations as the experiments showed promising results.


\newpage

\bibliography{ref}

\end{document}